\documentclass{article}

\PassOptionsToPackage{numbers}{natbib}
\usepackage[preprint]{neurips_2026}
\usepackage[utf8]{inputenc}
\usepackage[T1]{fontenc}
\usepackage{hyperref}
\usepackage{url}
\usepackage{booktabs}
\usepackage{amsfonts}
\usepackage{amsmath}
\usepackage{amssymb}
\usepackage{nicefrac}
\usepackage{microtype}
\usepackage{graphicx}
\usepackage{subcaption}
\usepackage{algorithm}
\usepackage{algorithmic}
\usepackage{enumitem}
\usepackage{xcolor}
\usepackage{wrapfig}
\usepackage{footnote}
\usepackage{fontawesome5}
\title{Bian Que: An Agentic Framework with Flexible Skill Arrangement for Online System Operations}

\author{
  \textbf{Bochao Liu}\textsuperscript{*},
  \textbf{Zhipeng Qian}\textsuperscript{*},
  \textbf{Yang Zhao}\textsuperscript{*},
  \textbf{Xinyuan Jiang}\textsuperscript{*},
  \textbf{Zihan Liang},\\
  \textbf{Yufei Ma},
  \textbf{Junpeng Zhuang},
  \textbf{Ben Chen}\textsuperscript{\dag},
  \textbf{Shuo Yang},
  \textbf{Hongen Wan},\\
  \textbf{Yao Wu},
  \textbf{Chenyi Lei},
  \textbf{Xiao Liang}\\
  \\
  Kuaishou Technology\\
   \faEnvelope\ \texttt{benchen4395@gmail.com}
}
\begin{document}

\maketitle
\renewcommand{\thefootnote}{}
\footnotetext{$^*$These authors contributed equally. $^\dag$The corresponding author.}
\renewcommand{\thefootnote}{\arabic{footnote}}
% ================================================================
% ABSTRACT
% ================================================================

\begin{abstract}
Operating and maintaining (O\&M) large-scale online engine systems (eg, search, recommendation and advertising) demands substantial human effort for release monitoring, alert response, and root cause analysis. 
Despite the inherent suitability of LLM-based agents for such operational scenarios, the critical bottleneck impeding their practical deployment lies not in reasoning, but in \emph{orchestration} capability—specifically, the precise selection of relevant \textit{data} (encompassing metrics, logs, and change events) and applicable \textit{knowledge} (including handbook-defined rules and empirically derived practitioner experience) tailored to each individual operational event.
Feeding all signals indiscriminately causes dilution and hallucination, while manually curating the event-to-(data, knowledge) mapping is intractable under dozens of daily releases. Here we present \textsc{Bian Que}, an agentic operating framework with three contributions: 
(i) The \emph{unified operational paradigm}, which abstracts routine daily O\&M actions into three canonical patterns: release interception, proactive inspection, and alert root cause analysis; 
(ii) The \emph{flexible Skill Arrangement}, each predefined Skill explicitly defines the requisite data and operational knowledge for each specific context. Such Skills can be automatically generated and updated by LLM agents, and can also be iteratively optimized by on-call engineers via natural language instructions.
(iii) The \emph{unified self-evolving mechanism}, where each correction signal enables two parallel evolutionary pathways: distilling event memory into knowledge, and targeted refinement of corresponding Skills.
Deployed on the e-commerce search engine of KuaiShou, \textsc{Bian Que} reduces alert volume by 75\%, achieves 80\% root-cause analysis accuracy, cuts mean time to resolution by over 50\%, and attains a 99.0\% pass rate on offline evaluations. Codes are available at \url{https://github.com/benchen4395/BianQue_Assistant}.
\end{abstract}

% ================================================================
% INTRODUCTION
% ================================================================

\section{Introduction}
\label{sec:introduction}

Large-scale online engine systems such as search, recommendation, and advertising platforms serve as the fundamental infrastructure of modern Internet services. The operation and maintenance (O\&M) of these systems consume considerable engineering resources. Furthermore, underlying modules and services iterate at a high frequency, with dozens of version releases deployed daily. Coping with such rapid iteration requires thorough comprehension of system internal mechanisms and continuous monitoring of runtime status, both of which impose heavy manual workloads. Fortunately, the advances in large language models (LLMs) and autonomous agents (Openclaw~\citep{openclaw2026}, Claude Code~\citep{claudecode2026}, Harness engineering~\citep{openai2026harnessengineering}) have enabled the automation of complex O\&M tasks, providing a scalable solution to reduce reliance on manual labor and lower overall operational costs.

To enable the principled deployment of LLM-based agents, we first abstract the heterogeneous O\&M scenarios into a collection of standardized recurring task paradigms. Although daily operational workflows vary across business domains and engineers, practical defensive maintenance actions inherently follow unified behavioral patterns. We summarize typical paradigms and formalize them as the \emph{three lines of defense}: release-period system monitoring, periodic proactive health inspection, and post-alert root cause diagnosis. Existing LLM-based O\&M agents~\citep{wang2024rcagent} mainly focus on the final post-alert troubleshooting, regarding alert triggering as the sole task entry point. By contrast, our framework advances automated release monitoring and proactive inspection \emph{prior to} alert occurrence, relegating alert-driven diagnosis to a fallback measure rather than the primary response.

All three lines of defense share an identical cognitive logic: operators perceive system runtime signals, apply professional operational knowledge, and conduct reasoning to derive diagnostic results or operational decisions. This renders LLM-based agentic frameworks naturally applicable to O\&M tasks. However, real-world operational scenarios are highly complex and heterogeneous. We define system runtime signals as \textit{data}, covering time-series metrics, structured logs, change events, traces and business indicators. We also define empirical operational experience for troubleshooting as \textit{knowledge}, including handbook rules, typical failure patterns, and module-specific behavioral norms. Directly feeding all raw data into LLMs hinders the model from capturing truly relevant information. Moreover, operational knowledge is usually undocumented, evolves with system iterations, and varies across business lines, modules and operational contexts. Hence, the core challenge is not reasoning capability, but the contextual selection of appropriate \textit{data} and \textit{knowledge} for each operational event. Such data and knowledge also need post-event incremental updates to accelerate the disposal of future similar incidents. Given the combinatorial complexity of business and scenarios, manually maintaining event-to-data-knowledge mappings is labor-intensive and infeasible.

%% Underlying all three lines of defense is a consistent cognitive structure: practitioners perceive environmental signals, apply their operational knowledge of system behavior, and reason toward a diagnosis or decision. This renders LLM-based agentic architectures a natural choice. In practice, however, operational scenarios are highly complex and heterogeneous. We refer to the runtime signals emitted by the system as \textit{data}, including time-series metrics, structured logs, change events, traces, and business indicators. We refer to the distilled operational experience that guides diagnosis as \textit{knowledge}, including handbook rules, known failure patterns, and module-specific behavioral norms. Feeding all available data wholesale into an LLM makes it difficult for the model to identify what is truly relevant, while knowledge is largely undocumented, shifts with every system iteration, and may draw on entirely different rules depending on the business line, affected module, and operational context. Therefore, the core challenge lies not in reasoning ability itself, but in how to select the right \textit{data} and \textit{knowledge} for each operational event, and place them in context. Meanwhile, it is important to ensure that these data and knowledge can be updated or revised after each event is completed, so as to facilitate the rapid handling of similar events in the future. The combinatorial complexity between business lines and scenarios makes it impossible to manually organize such mappings, as it would be an endless and arduous labor.

The above insights yield three key design principles. First, by abstracting O\&M workflows into the three lines of defense, we deploy a dedicated agent for each paradigm, individually responsible for release interception, proactive inspection, and alert root cause analysis. Second, given rapid system iteration and dynamic operational contexts, the mapping between operational event-to-(\textit{data}, \textit{knowledge}) requires continuous updating. maintaining it via natural language interfaces, instead of static configurations, ensures interpretability and sustainability. Third, practitioner feedback implicitly reflects the appropriateness of contextual assembly and the value of empirical knowledge to be preserved. A single feedback signal can therefore drive the parallel evolution of both event mappings and domain knowledge, avoiding the need for separate independent update pipelines.

These insights motivate \textsc{Bian Que}\footnote{Named after the legendary ancient Chinese physician Bian Que, who diagnosed and prevented illnesses prior to symptom onset—embodying our framework’s proactive philosophy.}, an agentic O\&M framework for online engine systems (Figure~\ref{fig:framework}). Its foundation is a \emph{unified operational paradigm}: the three lines of defense are formalized under a shared structure, each with a specialized agent (sharing the same LLM but differing in scenario knowledge and invoked Skills). Built atop this, \emph{Flexible Skill Arrangement} enables automatic generation and incremental refinement of event-to-(\textit{data}, \textit{knowledge}) mappings, with on-call engineers intervening via natural-language instructions. Closing the loop, a \emph{unified self-evolving mechanism} channels each feedback signal into two parallel pathways—memory-to-knowledge distillation and targeted Skill refinement—realizing co-evolution of the knowledge base and mappings through a single loop. We summarize our contributions as follows:

% These three insights give rise to \textsc{Bian Que}\footnote{Named after the legendary physician Bian Que in ancient Chinese history, renowned for diagnosing and preventing ailments before symptoms became apparent, mirroring the proactive philosophy of our framework.}, an agentic O\&M framework for online engine systems (Figure~\ref{fig:framework}). At its foundation, a \emph{unified operational paradigm} organizes the three lines of defense under a shared formulation with one specialized agent each; all agents share the same LLM and differ only in their scenario-level knowledge and the Skills they invoke. Built on top of this, \emph{Flexible Skill Arrangement} lets agents automatically generate and incrementally refine the event-to-(\textit{data}, \textit{knowledge}) mapping, with on-call engineers able to intervene through natural-language instructions. Closing the loop, a \emph{unified self-evolving mechanism} routes each feedback signal into two parallel pathways, memory-to-knowledge distillation and targeted Skill refinement, so that the knowledge base and the mapping co-evolve through a single loop. We summarize our contributions as follows:
\begin{itemize}[leftmargin=*]
\item We propose a \textbf{unified operational paradigm} that abstracts heterogeneous O\&M work into three canonical patterns: release interception, proactive inspection, and alert root-cause analysis.
\item We design \textbf{Flexible Skill Arrangement}, where each Skill specifies the relevant \textit{data} and \textit{knowledge} for each context and can be generated, updated, and corrected through natural-language feedback.
\item We introduce a \textbf{unified self-evolving mechanism} in which one feedback signal simultaneously drives memory-to-knowledge distillation and targeted Skill refinement, enabling the knowledge base and the context-assembly mapping to co-evolve.
\item We evaluate \textsc{Bian Que} both online and offline. In online deployment on the e-commerce search engine of a short-video platform serving hundreds of millions of users, \textsc{Bian Que} reduces alert volume by 75\%, achieves 80\% root-cause analysis accuracy, and cuts mean time to resolution by over 50\%. In offline evaluation on a curated benchmark of 104 cases, it attains a 99.0\% pass rate.
\end{itemize}

% \section{Related Work}
% \label{sec:related}

% ================================================================
% METHOD
% ================================================================

\section{Method}
\label{sec:method}

\subsection{Problem Formulation and Architectural Overview}
\label{sec:formulation}

The operational maintenance of large-scale online engine systems (search, recommendation, advertising) gives rise to a stream of \emph{operational events} $e$, including release deployments, alert triggers, and scheduled inspection requests. For each event, the agent produces a structured output $o$ that includes a decision (e.g., ``rollback'' or ``safe to proceed''), supporting evidence, and recommended follow-up actions.
As explained in \S\ref{sec:introduction}, we identify three canonical patterns---release interception, proactive inspection, and alert root cause analysis---that collectively cover the bulk of the operational workload. All three share a common computational structure:
\begin{equation}
\label{eq:paradigm}
    o = f_{\text{LLM}}\bigl(\mathcal{D}(e),\; \mathcal{K}(e),\; p_s\bigr),
\end{equation}
where $\mathcal{D}(e)$ denotes the \textit{data} retrieved for event $e$, $\mathcal{K}(e)$ denotes the applicable \textit{operational knowledge}, $p_s$ denotes the scenario-specific prompt composition, and $f_{\text{LLM}}$ means the large language model performing structured reasoning. The prompt $p_s$ is further decomposed into an agent-level and one or more Skill-level components, which are depicted in Eq.~\ref{eq:composition}. 

\begin{figure}[t]
\centering
\includegraphics[width=0.95\linewidth]{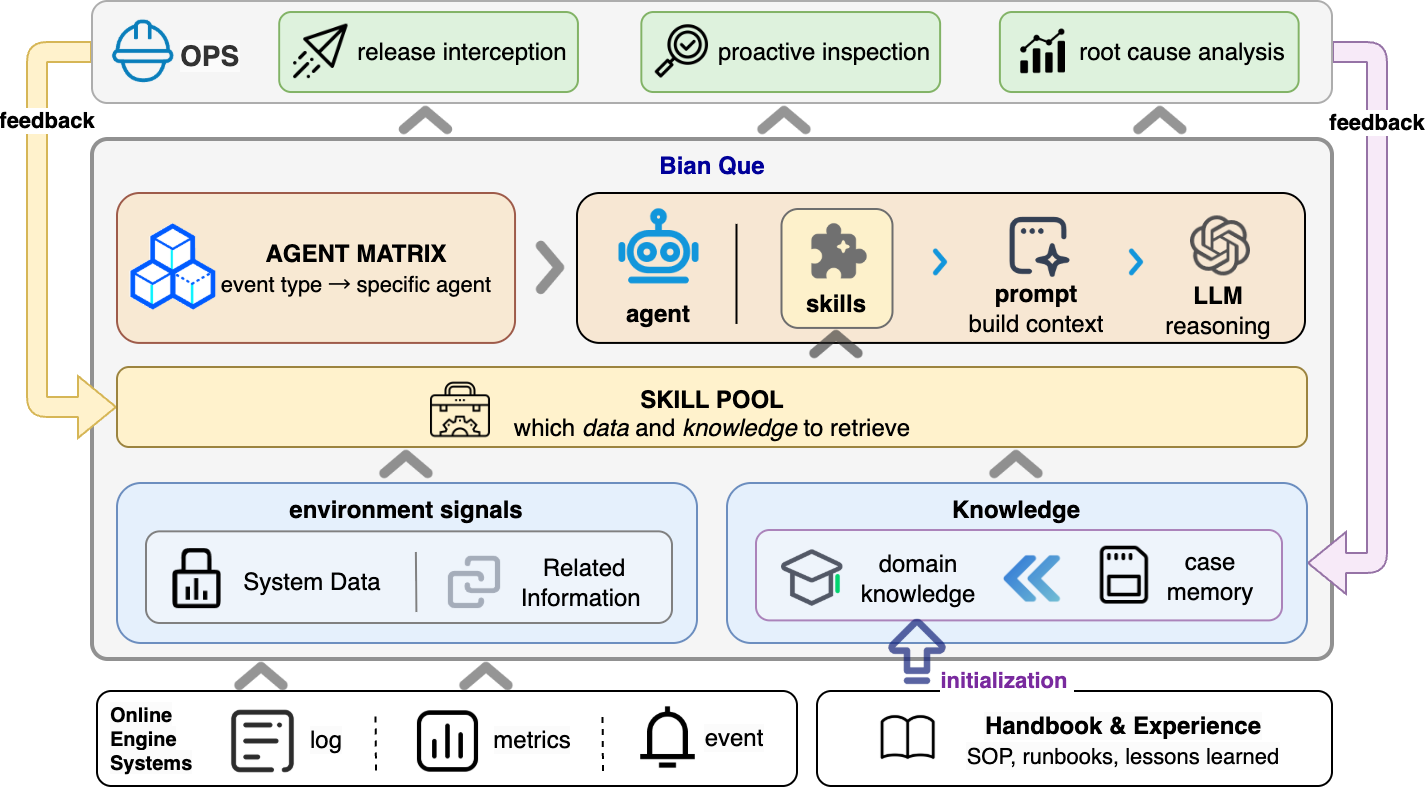}
\caption{Overview of the \textsc{Bian Que} architecture. Operational events from the OPS platform (top) are dispatched to a corresponding Agent, which invokes matched Skills to assemble the relevant \textit{data} (system signals: logs, metrics, change events) and \textit{knowledge} (domain knowledge distilled from case memory, seeded by operational handbooks). The LLM performs reasoning on the assembled information and returns diagnostic results. Practitioner feedback further drives two parallel self-evolution pathways (yellow: Skill refinement; purple: memory-to-knowledge distillation).}
\vspace{-0.3cm}
\label{fig:framework}
\end{figure}

Figure~\ref{fig:framework} shows the overall architecture. The three terms of Eq.~\ref{eq:paradigm} correspond exactly to the framework modules: the inherent data infrastructure of online engines $\mathcal{D}$, the two-stage knowledge base (\S\ref{sec:knowledge}) supplies $\mathcal{K}$, and a general-purpose LLM serves as $f_{\text{LLM}}$. 
Two key design features are highlighted as follows. First, instead of constructing dedicated monitoring infrastructure, this framework reuses the mature data stack of industrial online engines. The core challenge lies not in signal collection, but in \emph{selecting appropriate signals for individual events}, which is fulfilled by the flexible skill mechanism (Section \ref{sec:flexible_skill}). Second, the LLM is deployed as a pluggable reasoning module rather than a task-specific fine-tuned component. Empirical results (\S\ref{sec:model_comparison}) show that general-purpose models above roughly 30B scale perform comparably on operational reasoning when given adequate context, so the framework is model-agnostic above this scaling size.

\subsection{Agent Matrix and Skill Pool}
\label{sec:agent_matrix}
 
\begin{figure}[t]
\centering
\includegraphics[width=1\linewidth]{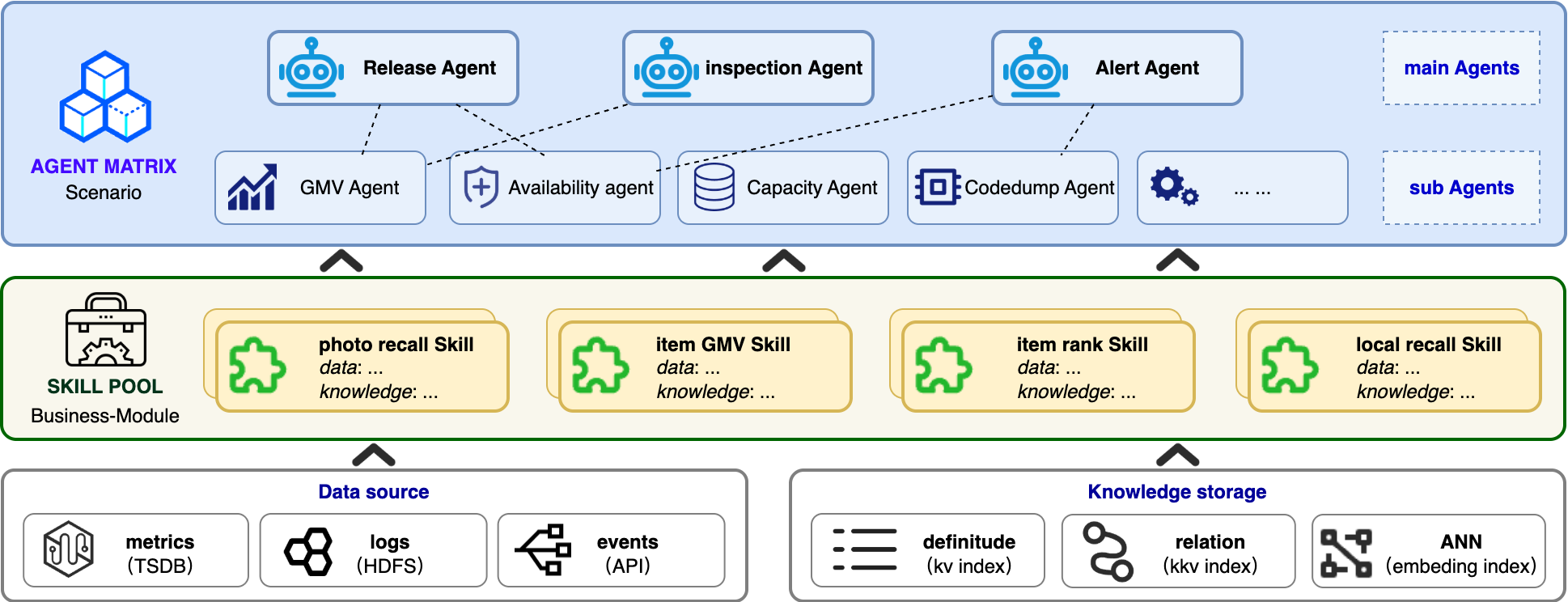}
\caption{Agent Matrix and Skill Pool. Each Agent (top) implements one canonical pattern; Skills (middle) are business-module orchestration units in a shared pool. Each single execution composes one Agent with matched Skills, which call into the knowledge base and data sources (bottom).}
\vspace{-0.5cm}
\label{fig:agent_matrix}
\end{figure}

A monolithic agent cannot efficiently handle the full breadth of online engine maintenance, because the data requirements and knowledge contexts vary drastically across business lines and service modules. We therefore adopt a matrix design that decomposes the problem along two orthogonal dimensions (shown in Figure~\ref{fig:agent_matrix}).

\textbf{Agents as Scenario Abstractions.}
Each agent type corresponds to one canonical pattern: the release interception agent implements checkpoint-based evaluation, the inspection agent implements periodic comprehensive analysis, and the alert agent implements event-triggered diagnosis. 
 
\textbf{Skills as Business-Module Orchestration Units.}
Each Skill specifies the \textit{data} and \textit{knowledge} needed for a particular combination of business and module. For instance, a Skill for ``recommendation recall module availability'' declares which metrics, logs, and change events to retrieve, which knowledge-base entries to consult (e.g., ``this module's GMV impact is typically small''), and how to reason over both. The internal structure of Skills is detailed in \S\ref{sec:flexible_skill}.
 
\textbf{Orthogonal Composition.}
A single agent execution composes the agent's scenario logic with one or more matched Skills:
\begin{equation}
\label{eq:composition}
    o = f_{\text{LLM}}\Bigl(\bigcup_{i} \mathcal{D}_{s_i}(e),\; \bigcup_{i} \mathcal{K}_{s_i}(e),\; p_{\text{agent}},\; \{p_{s_i}\}\Bigr), \quad \{s_i\} = \textsc{Match}(e, \mathcal{S}),
\end{equation}
where $\mathcal{S}$ is the Skill pool and $\textsc{Match}$ selects relevant Skills via keyword matching against event metadata. Here $p_s$ from Eq.~\ref{eq:paradigm} is realized as the composition of $p_{\text{agent}}$ with the Skill-level prompts $\{p_{s_i}\}$. This orthogonal decomposition allows agents and Skills to iterate independently: agents can improve reasoning strategies without modifying Skills, and Skills can evolve their data and knowledge configurations without changing agent logic.Within the inspection and alert categories, sub-agents are further specialized by metric family (e.g., GMV, capacity, availability, coredump) when a single agent prompt cannot cover the breadth of the category; sub-agents share the scenario logic of their parent and differ only in $p_{\text{agent}}$ specialization, and are therefore omitted from the formulation in Eq.~\ref{eq:composition}.
 
\subsection{Flexible Skill Arrangement}
\label{sec:flexible_skill}
 
The value of the Skill abstraction hinges on providing precise, scenario-appropriate context. Raw data tends to dilute critical signals and induce hallucinations, while overly rigid configurations may lead to information omission.
Considering the combinatorial complexity across business lines, modules and operational scenarios, as well as rapid system iteration, manual maintenance of Skill configuration becomes infeasible. The proposed Flexible Skill mechanism tackles this by enabling Skills to be \emph{LLM-generable, LLM-updatable}, and \emph{Human-correctable}.
 
\subsubsection{Skill Structure}
 
Each Skill is a structured document with three fields:
\begin{equation}
\label{eq:skill}
    \text{Skill} \;=\; \langle\, \texttt{LoadDataSchema},\; \texttt{Prompt},\; \texttt{Meta} \,\rangle.
\end{equation}
 
\texttt{LoadDataSchema} is a JSON specification that declares \emph{what to retrieve}. It lists the data-source calls (e.g., time-series metrics, structured logs, change events, traces) and knowledge-base queries to execute, along with their parameters and whether each source is mandatory or optional. \texttt{Prompt} is a structured template that defines \emph{how to reason}. It guides the LLM through a multi-step analysis over the retrieved data and knowledge to produce a structured diagnosis. At execution time, the Skill-level \texttt{Prompt} composes with the agent-level $p_{\text{agent}}$, and $p_{\text{agent}}$ supplies scenario logic (e.g., how to structure a release verdict), while the Skill \texttt{Prompt} supplies business-module-specific analysis steps. \texttt{Meta} contains name, version, description, and tags used for keyword-based Skill (\S\ref{sec:agent_matrix}).

% \begin{algorithm}[t]
% \caption{Flexible Skill Generation}
% \label{alg:skill_gen}
% \begin{algorithmic}[1]
% \REQUIRE Operational event $e$, source descriptor $L$, capability descriptor $C$, reference Skills $\mathcal{S}_{\text{ref}}$
% \ENSURE Generated Skill $s_{\text{cand}}$
% \STATE $s_{\text{cand}} \leftarrow f_{\text{LLM}}(e, L, C, \mathcal{S}_{\text{ref}})$
% \FOR{$\text{attempt} = 1$ \textbf{to} $3$}
%     \STATE $\text{data} \leftarrow \textsc{Execute}(s_{\text{cand}}.\mathtt{LoadDataSchema},\, e)$
%     \STATE $\text{errors} \leftarrow \textsc{Validate}(\text{data})$
%     \IF{$\text{errors} = \emptyset$}
%         \RETURN $s_{\text{cand}}$
%     \ELSE
%         \STATE $s_{\text{cand}} \leftarrow f_{\text{LLM}}(e, L, C, \mathcal{S}_{\text{ref}},\, \text{errors})$
%     \ENDIF
% \ENDFOR
% \RETURN $s_{\text{cand}}$
% \end{algorithmic}
% \end{algorithm}

\subsubsection{Skill Lifecycle}
\begin{figure}[t]
\centering
\includegraphics[width=1\linewidth]{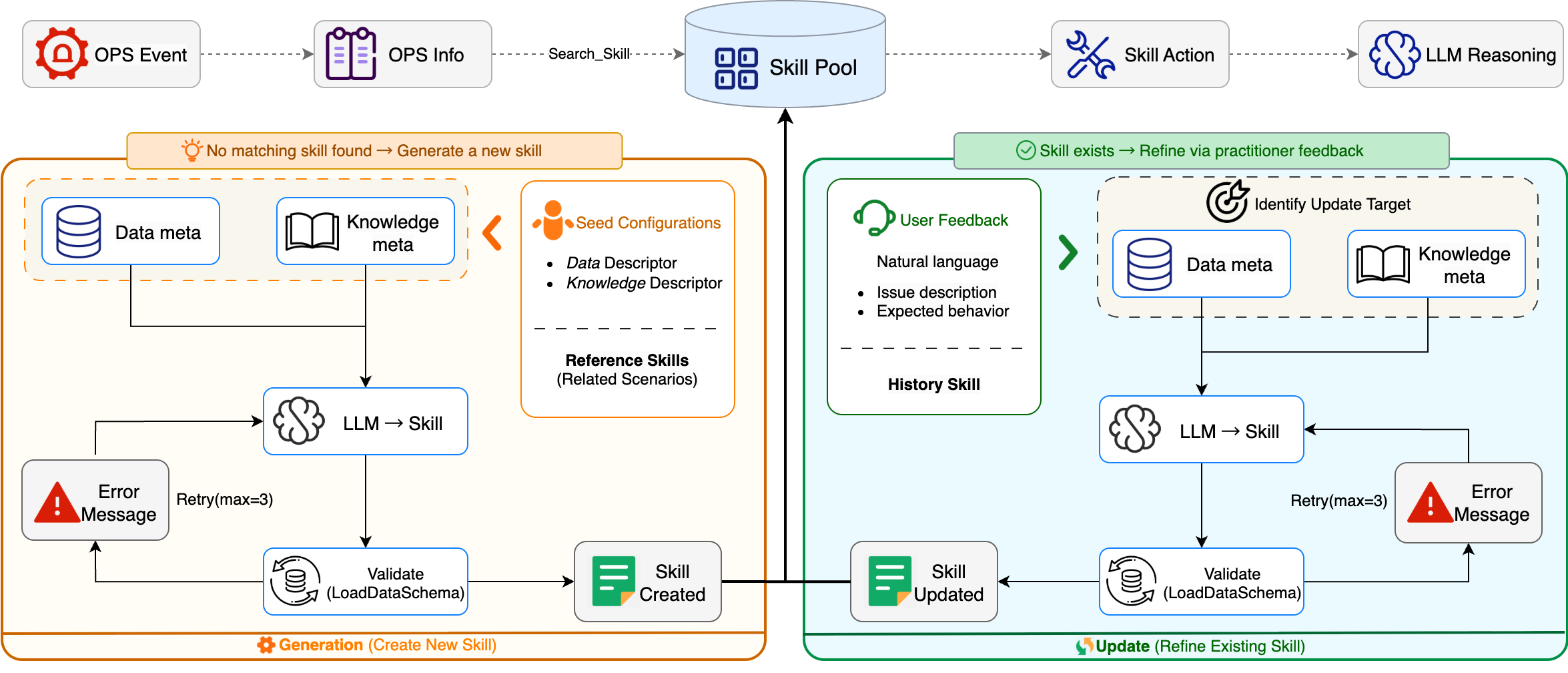}
\vspace{-0.6cm}
\caption{Flexible Skill lifecycle. New Skills are generated from seed configurations with validation and retry. Existing Skills are updated via practitioner feedback through a natural-language interface.}
\vspace{-0.6cm}
\label{fig:skill_lifecycle}
\end{figure}

The lifecycle consists of two phases (Figure~\ref{fig:skill_lifecycle}): \emph{Generation} creates a new Skill when no matching one exists for a given business-module combination, and \emph{Update} incrementally revises an existing Skill in response to practitioner feedback.

\textbf{Generation.}
When a new business-module combination encounters its first event and no matching Skill exists, the Skill framework triggers automated generation. The LLM takes two seed inputs: a \emph{source descriptor} that enumerates the available data sources and knowledge entries along with their semantics, and a \emph{capability descriptor} that specifies which of these sources are relevant to the scenario and how they relate to each other. The generated Skill is then validated by executing its \texttt{LoadDataSchema} against the triggering event; if any specified source fails to return the expected fields, the error is fed back to the LLM for regeneration, up to three retries. To facilitate cross-scenario transfer, existing Skills for related scenarios are provided as references during generation. 
% Algorithm~\ref{alg:skill_gen} gives pseudocode of the generation process.

\textbf{Update.}
Once deployed, a Skill may need to evolve as systems change. When a practitioner identifies a gap or an error in the agent's analysis, they describe the issue and expected behavior in natural language through a messaging interface. The framework then revises the appropriate component of the Skill---adjusting \texttt{LoadDataSchema} if the agent missed a critical data source or knowledge entry, or adjusting the \texttt{Prompt} if the agent retrieved adequate information but reasoned over it incorrectly. All cases follow the same validate-and-retry loop (up to three iterations) and require no code changes. 
 
% ============================================================
\subsection{Self-Evolving Knowledge and Skill via Unified Feedback}
\label{sec:knowledge}

A central design choice of \textsc{Bian Que} is that \emph{one} feedback signal simultaneously drives \emph{two} parallel learning pathways: one evolving the knowledge base, one evolving the Skills.
Every agent execution produces a case record $c = (e, \{s_i\}, \mathcal{D}, \mathcal{K}, o, f)$, where $f$ is the practitioner's feedback when available. The feedback is dispatched along two pathways:
\begin{equation}
\label{eq:feedback}
    f \;\longrightarrow\;
    \begin{cases}
        \mathrm{MemoryWrite}(c) \to \mathrm{KnowledgeDistill}(c) \\[3pt]
        \mathrm{SkillRefine}(\{s_i\},\, f)
    \end{cases}
\end{equation}
The knowledge pathway governs the accumulation of distilled operational experience, such as failure patterns and module-specific behavioral norms; the Skill pathway governs which \textit{data} and \textit{knowledge} the agent retrieves and how it reasons over them for each event.

\subsubsection{Knowledge Pathway}

The knowledge pathway operates in two phases. In the short-term phase, each case record is written into a memory store. Before each reasoning invocation, the agent retrieves the most relevant recent cases as ``working memory,'' enabling it to recognize recurring failure patterns within a recent window. In the long-term phase, each new case also triggers automatic knowledge extraction: the LLM distills reusable insights, such as causal relationships between modules, known failure patterns, and metric interpretation guidelines, into persistent domain knowledge entries. A daily offline process deduplicates and prunes the knowledge base to resolve accumulated contradictions and prevent unbounded growth; corrections take effect immediately through the short-term case memory and need not wait for this cycle. At deployment time, the knowledge base is additionally seeded with a small set of entries imported from operational handbooks, which follow the same indexing scheme and consolidation.
To support diverse retrieval needs, the knowledge base employs three complementary indices: a \textbf{KV index} for definitive knowledge keyed by business and scenario identifiers; a \textbf{KKV index} for relational knowledge, such as the impact of module~A on module~B for a given metric; and a \textbf{vector index} for fuzzy semantic retrieval. Skills address these indices through their \texttt{LoadDataSchema} (\S\ref{sec:flexible_skill}), coupling retrieval and knowledge at the query-specification level.

\subsubsection{Skill Pathway}
The same feedback signal is also analyzed to identify the root cause of any agent error and routed to the corresponding Skill lifecycle operation (\S\ref{sec:flexible_skill}). Two failure modes trigger a Skill \textbf{Update}: (1)~\emph{inadequate retrieval}, where a critical data source was absent or the wrong knowledge entry was queried, leading to a revision of \texttt{LoadDataSchema}; and (2)~\emph{flawed reasoning}, where the retrieved information was adequate but the \texttt{Prompt} combined it incorrectly, leading to a revision of the \texttt{Prompt}. Both operations follow the validation-with-retry procedure of \S\ref{sec:flexible_skill} and require no code changes. A third case, \emph{incorrect knowledge content}, arises when the retrieved entry itself is wrong or outdated; this falls outside the Skill pathway and is instead handled by the knowledge pathway: corrected knowledge takes effect immediately via short-term case memory, while the long-term entry is pruned or replaced at the next daily consolidation.
\subsubsection{Co-Evolution of Knowledge and Skills}
\label{sec:coevo}
The two pathways above are coupled by construction: a richer knowledge base provides more reference entries for newly generated or updated Skills, while higher-quality Skills produce cleaner case records for knowledge distillation. We qualitatively observe this compounding effect in deployment; a controlled quantitative measurement is left to future work, as it requires a longer horizon than our current six-month deployment permits.

% ================================================================
% EXPERIMENTS
% ================================================================

\section{Experiments}
\label{sec:experiments}

% We evaluate \textsc{Bian Que} along four dimensions: production deployment effectiveness (\S\ref{sec:exp_production}), the Flexible Skill mechanism (\S\ref{sec:exp_skill}), ablation of key components (\S\ref{sec:exp_ablation}), and sensitivity to the underlying LLM backbone (\S\ref{sec:model_comparison}).

% ============================================================

\subsection{Experimental Setup}
\label{sec:exp_setup}

\paragraph{Implementation Details.}
\textsc{Bian Que} is deployed online to operate the e-commerce search engine of the short-video platform, serving hundreds of millions of daily active users. The system comprises dozens of core service modules maintained by a cross-functional team of approximately 100 engineers, with an average of dozens of releases per day. The framework has been in continuous online operation for over six months.
The default LLM backbone is Qwen3.5-35B-FP8 deployed on a single NVIDIA Tesla L20 GPU. Skills are stored as structured YAML documents and indexed by keyword tags. The knowledge base uses Redis for KV and KKV indices and a dedicated ANN service for the vector index. The memory store retains a rolling window of recent cases with daily consolidation. All pass@$k$ experiments use temperature sampling at $T=0.3$; other experiments use greedy decoding unless otherwise noted. For each pass@$k$ evaluation, the Skill directory is reset to a clean seed state before the run, so that Skills from prior evaluation runs do not leak into the current run.

\paragraph{Datasets and Evaluation Metrics.}
Table~\ref{tab:paradigm} summarizes the three patterns and where each is evaluated.
We adopt four metrics. \textbf{Alert Volume Reduction} measures the relative decrease in alerts requiring human attention after deployment. \textbf{RCA Accuracy} is the proportion of root-cause analyses confirmed correct by practitioners. \textbf{MTTR} (Mean Time to Resolution) is the average time from alert trigger to issue resolution. \textbf{PASS@$k$} runs each event $k$ times and counts it as passed if at least one run produces a correct Skill yielding the ground-truth diagnosis.

\begin{table}[t]
\centering
\caption{The three canonical operational patterns and their evaluation in this paper.}
\label{tab:paradigm}
\small
\resizebox{\linewidth}{!}{%
\begin{tabular}{@{}llll@{}}
\toprule
\textbf{Pattern} & \textbf{Target Problem} & \textbf{Core Capability} & \textbf{Evaluated In} \\
\midrule
Release Interception & Incidents from releases & Single-event detection & Alert vol.\ reduction \\
Proactive Inspection & Latent system risks & Multi-event analysis & PASS@$k$  \\
Alert Root Cause Analysis & Post-alert diagnosis \& remediation & Correlation-based attribution & PASS@$k$; RCA acc.; MTTR \\
\bottomrule
\end{tabular}%
}
\vspace{-0.5cm}
\end{table}

% ============================================================
\subsection{Online Production Deployment Results}
\label{sec:exp_production}

Table~\ref{tab:production} reports the key operational metrics before and after \textsc{Bian Que} deployment over a six-month evaluation period. % 
Relative numbers are normalized to the pre-deployment baseline. The third row (absolute non-actionable alerts) is the compound of rows~1 and 2 and is the most directly interpretable measure of practitioner pager load.
% ; it is reported separately to avoid misreading rows~1 and 2 as independent effects.

\begin{table}[h]
\centering
\vspace{-0.5cm}
\caption{Online production deployment results over a six-month window, comparing the month preceding \textsc{Bian Que} rollout to the six-month post-rollout period.} % [M10, Q3]
\label{tab:production}
\small

\begin{tabular}{@{}lcc@{}}
\toprule
\textbf{Metric} & \textbf{Pre-deployment} & \textbf{Post-deployment} \\
\midrule
\multicolumn{3}{@{}l}{\emph{Alert load (upstream and downstream effects):}} \\
Fired alerts (relative)                              & 100\%   & 25\% \;($\downarrow$75\%) \\
Non-actionable ratio among fired alerts              & 80\%    & 15\%  \\
Non-actionable alerts, absolute (relative)           & 100\%   & $\sim$5\% \;($\downarrow$$\sim$95\%) \\ % TODO: replace with actual count if available
\midrule
\multicolumn{3}{@{}l}{\emph{Diagnostic response for fired alerts:}} \\
RCA accuracy                                         & ---     & 80\%  \\
Alerts resolved within 5 minutes                     & ---     & 95\% of cases \\
MTTR (relative)                                      & 100\%   & $<$50\% \;($\downarrow$50\%+) \\
\bottomrule
\end{tabular}
\vspace{-0.3cm}
\end{table}

\paragraph{Alert load reduction combines two mechanisms.}
The 75\% drop in fired alerts is primarily driven by release interception and proactive inspection, which resolve problems before they breach alert thresholds. The reduction in non-actionable ratio (80\% $\rightarrow$ 15\%) results from both upstream and downstream effects: for upstream, the first two lines of defense eliminate unstable releases and slowly-degrading services that previously cascaded into non-actionable alerts; for downstream, the alert RCA agent classifies and suppresses non-actionable alerts rather than paging practitioners. Since the two ratios share the ``fired alerts'' denominator and that denominator itself shrinks by 75\%, they are \emph{not} independent multipliers. The practically relevant quantity is the absolute non-actionable alert volume surfaced to practitioners, which drops to roughly 5\% of the baseline ($0.25 \times 0.15 / 0.80 \approx 0.047$), a roughly 95\% reduction in pager noise. We report this compound number in row~3 of Table~\ref{tab:production}.

\paragraph{Downstream diagnostic quality.}
For fired alerts that do warrant attention, the alert RCA agent returns a structured diagnosis within 5 minutes in 95\% of cases, with an 80\% RCA accuracy. This compresses MTTR by over 50\%, reflecting not only faster diagnosis but also the reduction in cognitive load on practitioners, who now receive evidence-backed recommendations rather than raw monitoring dashboards. The 80\% RCA accuracy, while demonstrating strong practical value, leaves room for improvement; our error analysis reveals that the majority of incorrect diagnoses stem from insufficient domain knowledge for newly deployed services, a limitation that the self-feedback mechanism (\S\ref{sec:knowledge}) is designed to address over time and that motivates the ablation studies in \S\ref{sec:exp_ablation}.

% ============================================================
\subsection{Flexible Skill Mechanism Evaluation}
\label{sec:exp_skill}

% The Flexible Skill mechanism (\S\ref{sec:flexible_skill}) has two interacting phases that we evaluate in turn: automated initialization from seed configurations, and human-in-the-loop correction of residual failures.

\paragraph{Skill Initialization.}
\label{sec:exp_skill_init}
For offline evaluation of the Flexible Skill mechanism, we curate 104 real operational events sampled from production logs—44 alerts and 60 inspections—each annotated with a ground-truth diagnosis by senior site reliability engineers (SREs).
Table~\ref{tab:skill_init} reports pass@$k$ for automated Skill generation across the 104-event dataset.
Two findings are worth highlighting. First, the overall pass@1 of 78.8\% shows that the majority of Skills are correctly generated on the first attempt, which validates the effectiveness of the seed configuration files (source descriptor and capability descriptor) in providing the LLM with enough scaffolding to produce usable Skills. Second, with a modest retry budget of five attempts, the end-to-end generation pipeline reaches pass@5 of 94.2\%, leaving only six cases unresolved. These residual failures concentrate on edge conditions that involve ambiguous causal attribution or novel business domain knowledge absent from the current Skill pool. They motivate the human-in-the-loop correction mechanism we evaluate next.

\begin{table}[t]
\centering
\caption{Skill initialization results (pass@$k$) on 104 real operational events. ``pass@$k$'' denotes the fraction of events for which at least one of $k$ independently sampled Skill generations yields the ground-truth diagnosis.} % [Table 3 caption]
\label{tab:skill_init}
\small
\begin{tabular}{@{}lcccccc@{}}
\toprule
\textbf{Scenario} & \textbf{Total} & \textbf{pass@1} & \textbf{pass@2} & \textbf{pass@3} & \textbf{pass@4} & \textbf{pass@5} \\
\midrule
Alert      & 44  & 31 (70.5\%) & 36 (81.8\%) & 40 (90.9\%) & 42 (95.5\%) & 42 (95.5\%) \\
Inspection & 60  & 51 (85.0\%) & 52 (86.7\%) & 54 (90.0\%) & 55 (91.7\%) & 56 (93.3\%) \\
\midrule
\textbf{Overall} & \textbf{104} & \textbf{82 (78.8\%)} & \textbf{88 (84.6\%)} & \textbf{94 (90.4\%)} & \textbf{97 (93.3\%)} & \textbf{98 (94.2\%)} \\ % [Q1]
\bottomrule
\end{tabular}
\vspace{-0.3cm}
\end{table}

\begin{table}[t]
\centering
\caption{Human-in-the-loop correction on the six cases that failed at pass@5 during initialization. Here ``pass@$k$'' denotes the number of correction rounds attempted.} % [M17]
\label{tab:skill_correct_standalone}
\small
\begin{tabular}{@{}lcccccc@{}}
\toprule
\textbf{Scenario} & \textbf{Total} & \textbf{pass@1} & \textbf{pass@2} & \textbf{pass@3} & \textbf{pass@4} & \textbf{pass@5} \\
\midrule
Alert      & 2 & 1 (50.0\%) & 1 (50.0\%)  & 2 (100.0\%) & 2 (100.0\%) & 2 (100.0\%) \\
Inspection & 4 & 2 (50.0\%) & 3 (75.0\%)  & 3 (75.0\%)  & 3 (75.0\%)  & 3 (75.0\%)  \\
\midrule
\textbf{Overall} & \textbf{6} & \textbf{3 (50.0\%)} & \textbf{4 (66.7\%)} & \textbf{5 (83.3\%)} & \textbf{5 (83.3\%)} & \textbf{5 (83.3\%)} \\
\bottomrule
\end{tabular}
\vspace{-0.3cm}
\end{table}

\begin{table}[!t]
\centering
\caption{End-to-end pass@$k$ on all 104 events. Every case is first processed through initialization (up to 5 generation attempts, reaching the 94.2\% pass@5 baseline in Table~\ref{tab:skill_init}); column pass@$k$ then reports the result after $k$ additional correction rounds applied to the still-failing cases.} % [M17]
\label{tab:skill_end2end}
\small
\begin{tabular}{@{}lcccccc@{}}
\toprule
\textbf{Scenario} & \textbf{Total} & \textbf{pass@1} & \textbf{pass@2} & \textbf{pass@3} & \textbf{pass@4} & \textbf{pass@5} \\
\midrule
Alert      & 44  & 43 (97.7\%)  & 43 (97.7\%)  & 44 (100.0\%) & 44 (100.0\%) & 44 (100.0\%) \\
Inspection & 60  & 58 (96.7\%)  & 59 (98.3\%)  & 59 (98.3\%)  & 59 (98.3\%)  & 59 (98.3\%)  \\
\midrule
\textbf{Overall} & \textbf{104} & \textbf{101 (97.1\%)} & \textbf{102 (98.1\%)} & \textbf{103 (99.0\%)} & \textbf{103 (99.0\%)} & \textbf{103 (99.0\%)} \\
\bottomrule
\end{tabular}
\vspace{-0.3cm}
\end{table}

\paragraph{Human-in-the-Loop Correction.}
\label{sec:exp_skill_correct}
We asked senior SREs to provide targeted natural-language corrections for the six cases that failed at pass@5 in Table~\ref{tab:skill_init} (2 alert and 4 inspection). 
Table~\ref{tab:skill_correct_standalone} reports pass@$k$
k on these six cases alone, proving the effectiveness of human correction and demonstrating its complementarity to the model's self-correction; Table~\ref{tab:skill_end2end} reports end-to-end pass@$k$ on all 104 events under the combination of model's self-skill-initialization followed by human correction on the failed cases, reflecting the overall system performance.
Two findings emerge. First, a single round of human feedback closes most of the remaining gap: overall pass@1 on all 104 events rises from 94.2\% to 97.1\%, confirming that targeted natural-language correction is a strong and efficient supervision signal. Second, the two scenario types converge at different rates: alert Skills reach 100\% by pass@3, while inspection Skills plateau at 98.3\% from pass@3 onward, as the residual failures involve higher problem complexity that domain-knowledge iteration alone cannot resolve.

% ============================================================
\subsection{Ablation Study}
\label{sec:exp_ablation}

We evaluate two ablated variants on the same 104-event dataset used in \S\ref{sec:exp_skill_init}, under the same pass@$k$ protocol: 1) \textsc{Static}: Skills are manually authored and fixed, disabling LLM-driven generation and update. This isolates the contribution of the Flexible Skill mechanism (\S\ref{sec:flexible_skill}). 2) \textsc{NoKnow}: No historical knowledge is retrieved during reasoning, disabling both short-term case-memory retrieval and long-term distilled-knowledge retrieval (\S\ref{sec:knowledge}).
Tables~\ref{tab:ablation_static} and~\ref{tab:ablation_noknow} report pass@$k$ for each variant. Both can be compared directly against the full-framework results in Table~\ref{tab:skill_init}.

\begin{table}[t]
\centering
\caption{Ablation on \textsc{Static} Skills, where Skills are manually authored without LLM-driven generation or updates, tested on the 104-event dataset.}
\label{tab:ablation_static}
\small
\begin{tabular}{@{}lcccccc@{}}
\toprule
\textbf{Scenario} & \textbf{Total} & \textbf{pass@1} & \textbf{pass@2} & \textbf{pass@3} & \textbf{pass@4} & \textbf{pass@5} \\
\midrule
Alert      & 44  & 26 (59.1\%) & 30 (68.1\%) & 33 (75.0\%) & 34 (77.2\%) & 36 (81.8\%) \\
Inspection & 60  & 42 (70.0\%) & 45 (75.0\%) & 46 (76.7\%) & 51 (85.0\%) & 51 (85.0\%) \\
\midrule
\textbf{Overall} & \textbf{104} & \textbf{67 (65.3\%)} & \textbf{75 (72.1\%)} & \textbf{79 (75.9\%)} & \textbf{85 (81.7\%)} & \textbf{87 (83.7\%)} \\
\bottomrule
\end{tabular}
\vspace{-0.3cm}
\end{table}

\begin{table}[h]
\centering
\caption{Ablation on \textsc{NoKnow}, where both case-memory and distilled-knowledge retrieval are disabled during reasoning, tested on the 104-event dataset.}
\label{tab:ablation_noknow}
\small
\begin{tabular}{@{}lcccccc@{}}
\toprule
\textbf{Scenario} & \textbf{Total} & \textbf{pass@1} & \textbf{pass@2} & \textbf{pass@3} & \textbf{pass@4} & \textbf{pass@5} \\
\midrule
Alert      & 44  & 28 (63.6\%) & 34 (77.3\%) & 37 (84.1\%) & 39 (88.6\%) & 40 (90.9\%) \\
Inspection & 60  & 46 (76.7\%) & 46 (76.7\%) & 48 (80.0\%) & 49 (81.7\%) & 50 (83.3\%) \\
\midrule
\textbf{Overall} & \textbf{104} & \textbf{74 (71.2\%)} & \textbf{80 (76.9\%)} & \textbf{85 (81.7\%)} & \textbf{88 (84.6\%)} & \textbf{90 (86.5\%)} \\
\bottomrule
\end{tabular}
\vspace{-0.3cm}
\end{table}

\paragraph{Flexible Skill is the most critical component.}
As shown in Table~\ref{tab:ablation_static}, replacing Flexible Skills with static configurations causes the largest drop: overall pass@5 falls from 94.2\% to 83.7\% ($-$10.5~pp), and pass@1 from 78.8\% to 65.3\% ($-$13.5~pp). Alert Skills suffer more than inspection Skills (pass@5 $-$13.7~pp vs.\ $-$8.3~pp), as alert root cause analysis involves more complex attribution logic that hand-authored rules struggle to cover. Crucially, 16.3\% of cases remain unresolved even at pass@5 (vs.\ 5.8\% for the full framework), indicating that static rules carry irreducible blind spots rather than a sampling deficit.

\paragraph{Historical knowledge is irreplaceable.}
As shown in Table~\ref{tab:ablation_noknow}, disabling knowledge retrieval lowers pass@5 from 94.2\% to 86.5\% ($-$7.7~pp), with a nearly identical drop at pass@1 (
$-$7.6~pp). Because the gap between pass@1 and pass@5 is preserved, the loss is systematic rather than stochastic—it \emph{cannot} be recovered by drawing more samples. Without historical knowledge of past incidents and service-specific signal patterns, the LLM falls back on generic priors, and additional retries merely resample from the same under-informed distribution.

\begin{wrapfigure}{r}{0.5\linewidth}
\centering
\vspace{-10pt}
\includegraphics[width=\linewidth]{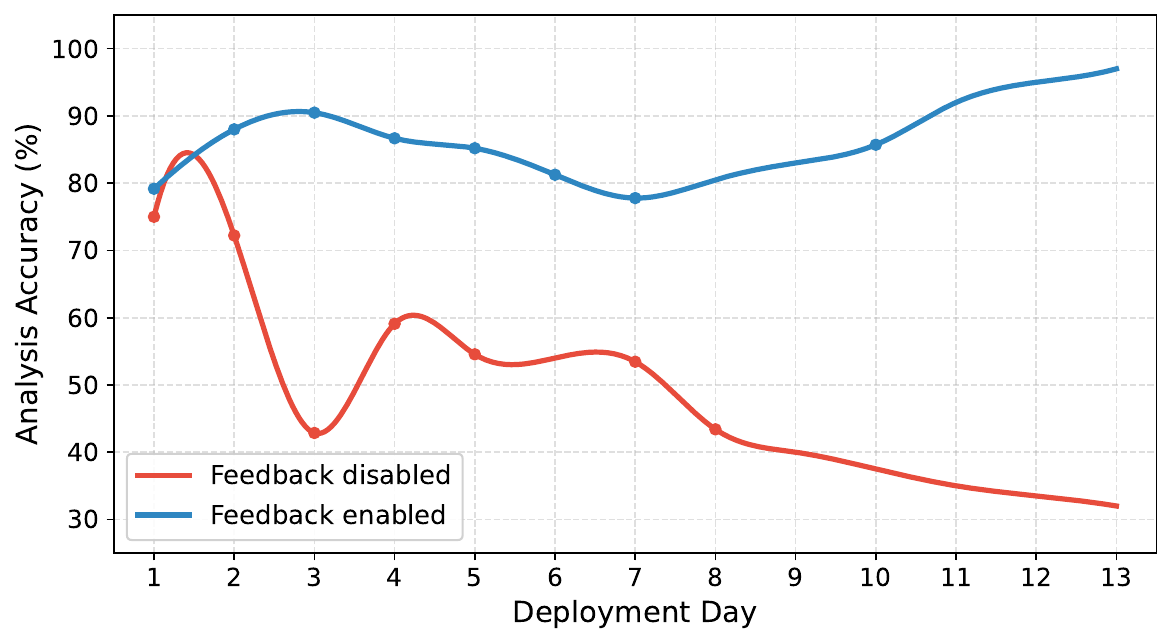}
\caption{Day-level alert-analysis accuracy on production traffic. Red: feedback disabled; Blue: feedback enabled.}
\label{fig:feedback_longitudinal}
\vspace{-8pt}
\end{wrapfigure}
\paragraph{Feedback-driven refinement sustains long-term accuracy.}

We provide two complementary pieces of evidence. \textbf{Offline:} as
 shown in Table~\ref{tab:skill_end2end}, on the 104-event dataset, a single round of practitioner feedback lifts overall pass@1 from 94.2\% to 97.1\%, and three rounds reach 99.0\%. \textbf{Online:} we deployed the feedback pathway on production traffic for a 13-day controlled experiment (Figure~\ref{fig:feedback_longitudinal}). The two arms begin from essentially the same initial state, with comparable day-1 accuracy ($\sim$75--80\%), but their trajectories diverge sharply thereafter. Without feedback, the system steadily loses coverage as production traffic drifts: newly emerging alert patterns fall outside the scope of the frozen Skill pool and knowledge base, and accuracy declines almost monotonically from $\sim$75\% to $\sim$32\% by day 13. With the feedback pathway enabled, the system continuously distills failure cases into new Skills and knowledge, and accuracy remains consistently above 80\% throughout the 13 days. The transient dip around day 7,triggered by a burst of novel failure modes,is absorbed within two days as the corresponding Skills are synthesized and deployed, indicating that the feedback loop \emph{actively repairs} coverage gaps rather than merely slowing degradation.

% ================================================================
% CONCLUSION
% ================================================================

\section{Conclusion}
\label{sec:conclusion}

We have presented \textsc{Bian Que}, an agentic framework for automated intelligent operations of large-scale online engine systems. The framework makes three contributions: a unified operational paradigm that provides complete yet lightweight coverage of the majority of maintenance tasks through three canonical patterns; a Flexible Skill Arrangement mechanism that jointly encapsulates \textit{data} routing and domain \textit{knowledge} and supports LLM-driven Generation and Update; and a unified feedback mechanism that simultaneously drives \textit{knowledge} base evolution and Skill refinement through practitioner signals. Production deployment on a large-scale search engine demonstrates substantial operational improvements: a 75\% reduction in fired alerts, a roughly 95\% reduction in practitioner-facing non-actionable alerts, 80\% RCA accuracy on the alerts that do require attention, and over 50\% MTTR compression.

% ================================================================
% REFERENCES
% ================================================================

\clearpage
\bibliographystyle{unsrtnat}
\bibliography{references}  % Uncomment when references.bib is ready
\appendix
\clearpage
\section{Related Work}
\label{sec:related}

\textbf{Intelligent Operations and AIOps}
Artificial Intelligence for IT Operations (AIOps)~\citep{zhang2024survey} is essential for large-scale online engine systems, empowering search, recommendation, and advertising, ensuring the normal operation of the system and algorithms.
Early AIOps research~\citep{notaro2021survey,chen2025stratus,riddell2026stalled,vitui2025empowering} addressed individual sub-problems such as anomaly detection~\citep{braei2020anomaly}, log parsing~\citep{he2017drain}, failure prediction~\citep{lin2018predicting}, and root cause localization~\citep{soldani2022anomaly} in isolation. More recently, LLM-based approaches have tackled log analysis~\citep{akhtar2025llm}, incident summarization~\citep{isaza2024retrieval}, conversational troubleshooting~\citep{roy2024exploring, pan2024raglog}, and agent-based diagnosis~\citep{wang2024rcagent, zhang2025survey}, validating the feasibility of applying LLMs to operational scenarios. However, these works share a common positioning: they assume a well-curated input context and focus on reasoning over it, rather than on assembling that context from raw production signals. \textsc{Bian Que} is positioned \emph{one step earlier} in the pipeline, responsible for selecting the right data and knowledge before any diagnostic reasoning happens. A specialized sub-task reasoner like RCAgent~\citep{wang2024rcagent} can, in principle, be plugged in as the $f_{\text{LLM}}$ component of our paradigm (Eq.~\ref{eq:paradigm}), consuming the $\mathcal{D}(e)$ and $\mathcal{K}(e)$ that our Flexible Skill layer prepares. Our distinctive contributions are accordingly (i) coverage of the operational \emph{lifecycle} under one unified paradigm, and (ii) treating the (\textit{data}, \textit{knowledge}) updating itself, based on both the reasoning and the existing knowledge, as an LLM-generated, continuously evolving artifact.

% \subsection{LLM-based Agents with Evolving Knowledge}
\textbf{LLM-based Agents with Evolving Knowledge}
LLM-based autonomous agents have established core paradigms for tool use and multi-step reasoning~\citep{yao2022react, schick2023toolformer, yang2023auto, shen2023hugginggpt, wang2023voyager, qiao2023taskweaver,liang2026ig}. More recent systems such as LangChain~\citep{langchain2026anatomyagentharness}, Claude Code~\citep{claudecode2026} and OpenClaw~\citep{openclaw2026} further demonstrate how agentic architectures can leverage standardized tool protocols~\citep{wang2025mcp} to operate over large-scale heterogeneous tool ecosystems. These frameworks generally assume that the relevant tools and data sources are either pre-specified or discoverable through exploration, an assumption that breaks down in industrial operations where heterogeneous data sources number in the thousands and the right subset varies drastically across business lines, modules, and scenarios~\citep{jiang2025llma4itops, yang2026aoi}. Our Flexible Skill mechanism addresses this by introducing an evolvable abstraction layer in which the data-routing logic itself is LLM-generated and incrementally updated through natural-language interfaces.
On the knowledge side, prior work spans RAG~\citep{lewis2020retrieval} and its variants~\citep{asai2023self, jeong2024adaptive}, memory-augmented architectures~\citep{zhong2024memorybank, shinn2023reflexion, park2023generative, xu2025mem}, and self-refinement methods~\citep{madaan2023self, tao2024survey}. Recent work has further explored self-evolving memory systems that learn to update their own memory operations from feedback~\citep{zhang2026memskill, zhang2026live, wei2025evo}. However, conventional RAG treats the knowledge base as static, and most self-evolving mechanisms operate along a single feedback pathway---improving either the knowledge base or the agent behavior, but not both. Our framework introduces a unified feedback mechanism in which each correction signal simultaneously drives memory-to-knowledge distillation and Skill refinement, enabling the two to co-evolve through a shared loop.

\begin{table}[h]
\centering
\caption{Comparison of LLM scaling on root-cause analysis (of 36 events). pass@$k$ is reported as correct cases (\%).}
\label{tab:backbone}
\small
\begin{tabular}{@{}lccccc@{}}
\toprule
\textbf{Model} & \textbf{pass@1} & \textbf{pass@2} & \textbf{pass@3} & \textbf{pass@4} & \textbf{pass@5} \\
\midrule
GLM5                 & 28 (77.8\%) & 30 (83.3\%) & 32 (88.9\%) & 33 (91.7\%) & 35 (97.2\%)  \\
DeepSeek-V3.2        & 28 (77.8\%) & 31 (86.1\%) & 33 (91.7\%) & 35 (97.2\%) & 36 (100.0\%) \\
Qwen3.5-35B-FP8      & 26 (72.2\%) & 29 (80.6\%) & 32 (88.9\%) & 34 (94.4\%) & 34 (94.4\%)  \\
Qwen3.5-27B-FP8      & 23 (63.9\%) & 25 (69.4\%) & 29 (80.6\%) & 30 (83.3\%) & 32 (88.9\%)  \\
Qwen3.5-9B           & 17 (47.2\%) & 22 (61.1\%) & 26 (72.2\%) & 28 (77.8\%) & 29 (80.6\%)  \\
Qwen3.5-4B           & 13 (36.1\%) & 15 (41.7\%) & 18 (50.0\%) & 20 (55.6\%) & 21 (58.3\%)  \\
Qwen3.5-0.8B         & 5  (13.9\%) & 6  (16.7\%) & 8  (22.2\%) & 8  (22.2\%) & 9  (25.0\%)  \\
\bottomrule
\end{tabular}
\vspace{-0.5cm}
\end{table}
\section{LLM Backbone Comparison.}
\label{sec:model_comparison}
To evaluate sensitivity to the underlying LLM, we compare seven models (of various versions and scalings) on the alert root-cause analysis task with a 36-event subset described in \S\ref{sec:exp_setup}. Before each evaluation run, the Skill directory is reset to the clean seed state to eliminate cross-run contamination. All models use temperature $T=0.3$. We find that, frontier models cluster tightly at the top and clearly separate from smaller ones: DeepSeek-V3.2, GLM5, and Qwen3.5-35B achieve pass@1 within a 5.6-point band (77.8\%, 77.8\%, and 72.2\% respectively); DeepSeek-V3.2 is the only model reaching 100\% at pass@5. While dropping from 35B to 9B costs 25\% of pass@1 (72.2\% $\rightarrow$ 47.2\%), and further reductions to 4B and 0.8B cause catastrophic degradation (36.1\% and 13.9\%). Practical recommendation is to use a backbone of around 35B parameters or larger for online deployment of \textsc{Bian Que}; smaller models exhibit substantially less robust behavior on operational reasoning tasks.

\section{Limitations and Future Work.}
% \paragraph{Generality.}
Several directions remain open. First, the current framework covers the reasoning and diagnosis stages of operational workflows but does not automate the subsequent execution of remediation actions (e.g., performing rollbacks or scaling operations); extending the framework to closed-loop autonomous remediation is a natural next step. Second, while the Agent Matrix currently employs keyword-based Skill matching, more sophisticated routing mechanisms (e.g., learned embeddings over event metadata) could improve matching accuracy for novel event types. Third, the coordination across multiple agent executions within a single incident remains largely manual; developing orchestration protocols for multi-agent collaboration in complex, cascading failure scenarios presents an interesting research direction. Fourth, directly quantifying the compounding effect between the knowledge pathway and the Skill pathway requires a longer evaluation horizon than our current six-month deployment permits (\S\ref{sec:coevo}) and is left to future work. Finally, the residual RCA errors concentrate on newly deployed services for which the knowledge base has not yet accumulated sufficient experience; we expect the self-feedback mechanism to narrow this gap over time, but validating that hypothesis will require a multi-year evaluation.

% \newpage
% \input{checklist.tex}

\end{document}